\pdfoutput=1

\documentclass[11pt]{article}

\usepackage[final]{acl}

\usepackage{times}
\usepackage{latexsym}

\usepackage[T1]{fontenc}

\usepackage[utf8]{inputenc}

\usepackage{microtype}

\usepackage{inconsolata}

\usepackage{graphicx}
\usepackage{amsmath}
\usepackage{amsfonts}
\usepackage{microtype}
\usepackage{booktabs} 
\usepackage{xspace}
\usepackage{subcaption}
\usepackage{bbm}
\usepackage{multirow}
\usepackage{enumitem}
\usepackage{xspace}
\usepackage{cleveref}
\usepackage{adjustbox}
\usepackage{tikz}
\usepackage{tabularx}
\usepackage{xcolor}
\usepackage{fancyvrb}
\usepackage{fvextra}

\DeclareMathOperator*{\argmax}{arg\,max}

\usetikzlibrary{trees}
\newcommand{\draftonly}[1]{#1} 
\renewcommand{\draftonly}[1]{}

\newcommand{\bleu}{\textsc{Bleu}\xspace}
\newcommand{\rouge}{\textsc{Rouge}\xspace}
\newcommand{\meteor}{\textsc{Meteor}\xspace}
\newcommand{\bertscore}{\textsc{BertScore}\xspace}
\newcommand{\sgd}{\textsc{SGD}\xspace}
\newcommand{\abcd}{\textsc{ABCD}\xspace}

\usepackage{xspace}
\usepackage{cleveref}
\newcommand{\Scref}[1]{\S\ref{#1}}
\usepackage{tcolorbox}

\title{Unsupervised Extraction of Dialogue Policies from Conversations}

\author{Makesh Narsimhan Sreedhar, Traian Rebedea \and Christopher Parisien\\
  NVIDIA \\
  Santa Clara, CA \\
  \texttt{\{makeshn, trebedea, cparisien\}@nvidia.com} 
  }

\begin{document}
\maketitle

\begin{abstract}
Dialogue policies play a crucial role in developing task-oriented dialogue systems, yet their development and maintenance are challenging and typically require substantial effort from experts in dialogue modeling. While in many situations, large amounts of conversational data are available for the task at hand, people lack an effective solution able to extract dialogue policies from this data.  In this paper, we address this gap by first illustrating how Large Language Models (LLMs) can be instrumental in extracting dialogue policies from datasets, through the conversion of conversations into a unified intermediate representation consisting of canonical forms. We then propose a novel method for generating dialogue policies utilizing a controllable and interpretable graph-based methodology. By combining canonical forms across conversations into a flow network, we find that running graph traversal algorithms helps in extracting dialogue flows. These flows are a better representation of the underlying interactions than flows extracted by prompting LLMs. Our technique focuses on giving conversation designers greater control, offering a productivity tool to improve the process of developing dialogue policies.
\end{abstract}

\section{Introduction}
\label{sec:introduction}

Chatbots and virtual assistants have emerged as powerful tools for guiding users or automating specific tasks across different domains, from facilitating restaurant reservations~\cite{budzianowski-etal-2018-multiwoz} to handling product returns on e-commerce platforms~\cite{chen-etal-2021-action}. 



Most task-oriented dialogue systems (TODS) nowadays use two key components: a Natural Language Understanding (NLU) engine and a Dialogue Manager~\cite{bocklisch2017rasa}. The role of the NLU engine is to perform intent detection and slot extraction, essential for understanding the user's requests. Concurrently, the Dialogue Manager leverages the current dialogue state, alongside the intent and slots identified from the latest user message, to determine the subsequent bot action or response. In most cases, both the NLU and Dialogue Manager rely on expert human intervention, typically involving a mix of conversation designers and data scientists. The NLU component requires a predefined set of user intents and slots 
while the Dialogue Manager necessitates 
dialogue policies that dictate the bot responses.

In the development of task-oriented assistants, it is common to have access to a corpus of pre-existing conversations.
Recent research has shown considerable interest in harnessing these conversational corpora to construct TODS. Extracting intents directly from these dialogues has 
demonstrated significant potential in augmenting the efficiency of conversation designers in modeling the NLU component~\cite{chatterjee2020intent, kumar-etal-2022-intent, du-etal-2023-two}. However, the task of deriving dialogue policies from the same set of conversations presents a more complex challenge and requires a nuanced understanding of conversational dynamics and objectives. Only a limited number of studies have ventured into this domain, exploring methodologies for automatic dialogue policy extraction~\cite{richetti2017analysis, vakulenko2019qrfa, ferreira-2023-automatic}.

In this paper, we introduce a novel hybrid methodology (\Scref{sec:method}) that combines Large Language Models (LLMs) with graph-based algorithms for the automated extraction of dialogue policies from a corpus of task-specific conversations. To that end, we first translate the turns in each dialogue into canonical forms~\cite{sreedhar-parisien-2022-prompt} using an LLM. The canonical forms are then clustered together to smooth out minor variations, following which we construct a graph modelling the entire corpus of conversations. This graph is akin to a \textit{flow network}, where nodes represent canonical forms of dialogue turns and edges signify the progression and connection between different turns. Finally, we apply path-finding algorithms to this graph to extract dialogue policies.

The proposed approach combines sequences of user and assistant canonical forms, that can be seen as \textit{dialogue trajectories}, into a more complex \textbf{\textit{dialogue policy}}. Thus, the extracted policies can handle digressions that are expressed using a branching logic determined by the intent of a user message. As all trajectories in our policies, including digressions, are composed of sequences of (user and bot) canonical forms, we can express the dialogue policy extraction from a corpus of conversations as a compositional task that combines translation and multi-document summarization. This enables using automatic metrics (e.g. \bleu, \bertscore) for evaluating the quality of the generated policies (\Scref{sec:experiments}). We also show that these metrics correlate very well with human evaluation (\Scref{sec:analysis}). 

Our findings indicate superior performance of our hybrid graph and LLM-based approach over techniques that rely solely on prompting LLMs for policy generation. In addition to better quantitative performance, the graph-based methodology provides enhanced controllability, interpretability, and robustness. These qualities render it a practical and effective tool for aiding conversation designers, in contrast to the more opaque, black-box nature of prompt-based LLM approaches.

Our main contributions are as follows:
\begin{itemize}[noitemsep,topsep=0pt]
    \item Demonstrating the feasibility of extracting dialogue policies expressed as sequences of user and assistant canonical forms from a corpus of conversations focused on a specific task.
    
    \item Modelling conversations with a flow network graph derived from the sequences of canonical forms provides an efficient method for policy extraction. The evaluation of dialogue policies computed using our hybrid graph and LLM approach demonstrates superior performance compared to prompt-based methods.

    \item Providing a controllable and  highly interpretable practical solution 
    to be used by conversation designers in real-world scenarios. 

    \item Contributing to the field by releasing the extracted dialogue policies for tasks in two popular datasets for TODS: SGD~\cite{rastogi2020towards}  and ABCD~\cite{chen-etal-2021-action}.
\end{itemize}

\section{Background}
\label{sec:background}

\paragraph{Task-Oriented Dialogue.} Most tools for building task-oriented chatbots and virtual assistants use two different components: NLU and a Dialogue Manager~\cite{liu2021benchmarking}. These range from commercial solutions (e.g. Google DialogFlow~\cite{dialogflow} or Oracle Digital Assistant~\cite{bors2020oracle}) to open-source tools like Rasa~\cite{bocklisch2017rasa} or research-focused platforms such as ConvLab~\cite{lee-etal-2019-convlab}. 


\begin{figure*}
    \centering
    \includegraphics[width=\textwidth]{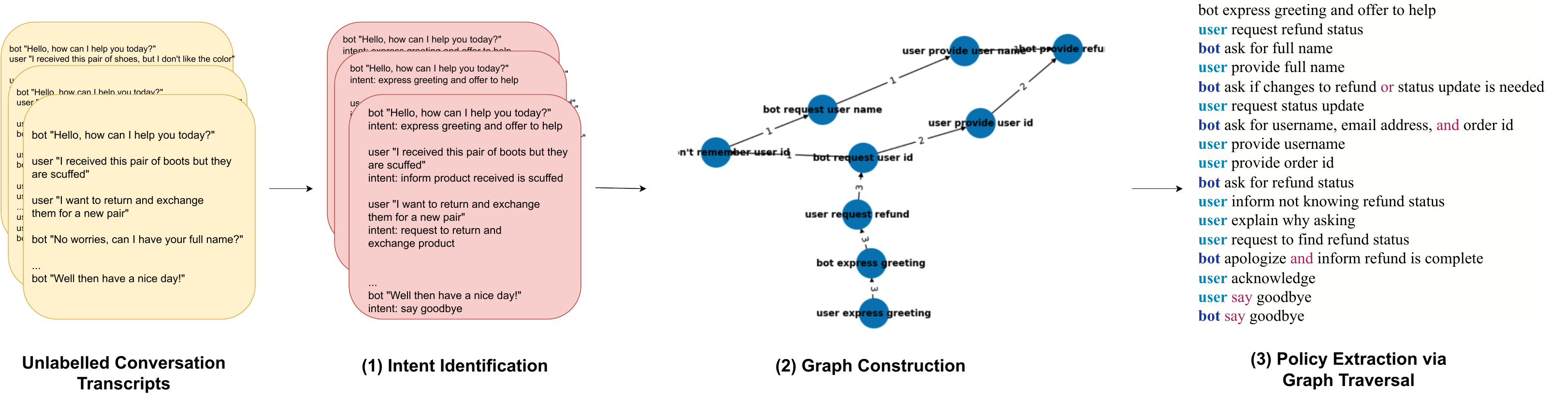}
    \caption{The three stages of the proposed solution for extracting dialogue flows: 1) Label user and bot turns in the conversations with canonical forms (\Scref{sec:intent_identification}); 2) Construct an interaction graph between user and bot canonical forms (\Scref{sec:graph_construction}); 3) Use graph traversal to extract dialogue flows as sequences of canonical forms (\Scref{sec:policy_extraction}).} 
    \label{fig:proposed-method-pipeline}
\end{figure*}

Dialogue policies can be modeled as sequences of user intents and bot actions, for example using \textit{stories} in Rasa~\cite{bocklisch2017rasa} or \textit{Colang flows} in NeMo Guardrails~\cite{rebedea2023nemo}. Our work is valuable for this modeling: the extracted dialogue policies can serve as starting points for conversation designers to refine. 

Traditionally, the development of task-oriented dialogue systems (TODS) required manual effort from experts. However, recent tools and methods aim to reduce this effort by leveraging large datasets for automatic intent discovery, with some addressing the challenge of dialogue policy generation. Even end-to-end neural TODS~\cite{hosseini2020simple} that embed intents and policies in model weights can use the extracted human-readable dialogue policies to enhance the explainability of the underlying opaque systems.

\paragraph{Canonical Forms.} NLU has typically used discriminative components for intent classification and slot labeling. With advances in generative text models~\cite{radford2018improving, brown2020language}, NLU can now be remodeled as a generative engine for intents and slots. Canonical forms~\cite{sreedhar-parisien-2022-prompt} encode the intent of conversation turns in a concise, standard form. Unlike the closed set of expert-defined intent classes, canonical forms are generated by models and are task-independent, offering a flexible way to encode dialogue policies~\cite{rebedea2023nemo}.

\paragraph{Intent Discovery.}
Intent mining has lately been an active topic not only in conversations~\cite{chatterjee2020intent}, but also in web queries~\cite{vedula2020open}. Most of the works employ various clustering algorithms (DBSCAN~\cite{chatterjee2020intent}, $k$-means~\cite{du-etal-2023-two}, iterative~\cite{benayas2023automated}), with different text embeddings. Recent works propose using contrastive learning for training specific embeddings for this task~\cite{du-etal-2023-two, kumar-etal-2022-intent} or using a dual-stage clustering~\cite{du-etal-2023-two}. Similar to our proposed intent discovery approach, some methods generate intent names as well~\cite{vedula2020open, benayas2023automated}. Our intent discovery method combines an LLM p-tuned for generating canonical forms with an extra clustering step and is applied for both user and bot intents. 

\paragraph{Dialog Flow Extraction.}
There are just a handful of works tackling the generation of dialogue policies. Earlier works employed a type of process mining that required either using a taxonomy of speech acts~\cite{vakulenko2019qrfa} or other predefined classes~\cite{richetti2017analysis} for each turn in the conversation. \citet{ferreira-2023-automatic} is the most similar to our proposed method, as it considers a graph-based approach to identify frequent sequences of turn types, but it employs a specific taxonomy of dialogue acts to label the turns. One important advantage of our approach is that it does not need any human intervention. 

Unsupervised dialogue structure discovery~\cite{lu-etal-2022-unsupervised, shi-etal-2019-unsupervised} is similar to dialogue flow extraction. However, there are important differences: they mainly aim to discriminate conversations in different tasks from a dataset and the latent structures used to encode the state of a conversation cannot be easily used by the Dialogue Manager of a TODS. Another task that has some similarities is workflow discovery which aims at predicting API calls given a task-oriented conversation, but it was only explored in a low-data regime, not fully unsupervised~\cite{hattami2022workflow}.

\section{Method}
\label{sec:method}
To extract dialogue policies (flows) from conversational data, we propose a pipeline comprising three key stages: intent identification, graph construction utilizing the identified intents, and the application of graph traversal algorithms for the extraction of dialogue flows. The functionality is depicted in Fig.~\ref{fig:proposed-method-pipeline}:
the input is a corpus of conversations on a given task and the output is the dialogue policy as a combination of sequences of canonical forms.

\subsection{Intent Identification}
\label{sec:intent_identification}
We begin with a corpus of task-specific conversations, such as customer interactions regarding product returns in an e-commerce setting. These conversations are structured as a series of exchanges between an user and a human agent, composed of $n$ dyads. The typical format of a conversation is an alternating sequence of user and agent turns, represented as $[u_1, a_1, ..., u_n, a_n ]$.

The primary objective at this initial stage of the pipeline is to assign an intent label to each turn in the conversation, effectively mapping:
$\text{turn}_{i} \rightarrow \text{intent}(\text{turn}_{i})$
This enables us to analyze conversations at a higher level of abstraction rather than operating at the level of individual turns. The abstracted conversation can thus be depicted using the corresponding intents, $[\text{intent}(u_1), \text{intent}(a_1) ..., \text{intent}(a_n) ]$.

Not only that intent labels are not available in an unsupervised context, but the intents provided as part of TOD datasets are not easily transferable across domains. We adopt the usage of \textit{canonical forms}~\cite{sreedhar-parisien-2022-prompt} for inferring the intents of conversation turns. This approach offers a practical alternative, enabling intent identification without relying on predefined label sets.

\paragraph{Canonical Forms.} Intent labels traditionally tend to be terse, and this often hinders the generalization of models to new domains. Canonical forms are concise, yet descriptive phrases that can capture the essence of utterances in the conversation (see Fig.~\ref{fig:intent-detection}). More complex examples are shown in Appendix~\ref{sec:appendix:examples-flows} highlighting that canonical forms can also capture slots in addition to the intent (e.g. \texttt{bot ask for city}, \texttt{user provide city}).



\paragraph{Weak Supervision.} We leverage the impressive generalization capabilities of language models to extract canonical forms across a wide range of domains. Starting from a small set of 200 conversations from two tasks in the ABCD dataset (product returns and shipping inquiries), canonical forms for each turn are obtained using \texttt{text-davinci-003}, OpenAI's instruction-tuned LLM ~\cite{ouyang2022training}. Using this weakly supervised data, a smaller LLM(\Scref{sec:ptuned_llm}) is p-tuned~\cite{liu-etal-2022-p} to predict the canonical form for a particular turn given the conversation history, i.e. it learns the mapping: 
$[u_1, a_1, u_2, ..., u_i] \rightarrow \text{intent}(u_i)$.
The trained model is then used to annotate utterances with canonical forms across different domains. It is employed for all our experiments, showing its generalization not only to the other tasks in the ABCD dataset, but also to a different domain (SGD). 

Aligning a separate model allows for obtaining more consistent and cheaper annotations than using OpenAI models. Moreover, as the p-tuning dataset is small we also plan to obtain human annotations and release a commercially viable model for generating canonical forms. More details about the p-tuned model and prompting used for obtaining weak labels are shown in Appendix~\ref{sec:appendix:intent_identification}.

\begin{figure*}
    \centering
    \includegraphics[width=.9\textwidth]{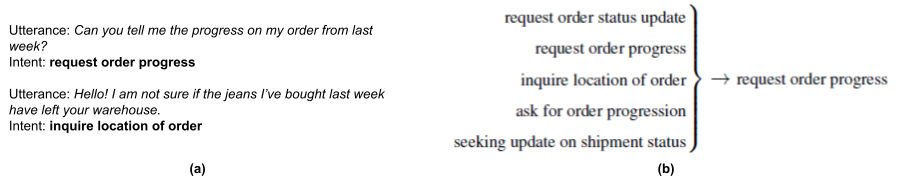}
    \caption{Intent identification: (a) Extraction of canonical forms from conversation turns using an LLM, (b) Intent normalization via clustering.}
    \label{fig:intent-detection}
\end{figure*}


\paragraph{Intent Normalization.} Using a generative approach to obtain the canonical form for utterances introduces variability in how similar intents are described (see Fig.~\ref{fig:intent-detection}). In this stage, we want to group canonical forms that represent identical intents, identify a representative form within each group, and subsequently re-annotate the conversations with the representative forms for each group. 


We extract the canonical forms not only for user turns, but also for agent responses. For the normalization stage, we use agglomerative clustering independently for the set of  canonical forms (user and agent). The embeddings for the canonical forms are computed using the \texttt{MiniLM-L6} model~\cite{reimers-gurevych-2019-sentence}. 
The representative canonical form for each cluster is chosen based on frequency.
All other canonical forms within each cluster are then substituted with this representative canonical form. This procedure yields a collection of conversations labelled with a consistent and unified set of canonical forms. 
Additional implementation details are in Appendix~\ref{sec:appendix:intent_identification}. 

\subsection{Graph Construction}
\label{sec:graph_construction}
The conversations with the unified canonical forms allow us to construct a graph denoting how each conversation proceeds. Let us consider a conversation with canonical forms $[\bar{u}_1, \bar{a}_1, \bar{u}_2, ... \bar{u}_n, \bar{a}_n]$ where $\bar{u}_i$ and $\bar{a}_i$ denote the canonical forms for user turn $u_i$ and agent turn $a_i$ respectively. This allows us to construct a linear path that denotes how the conversation progressed: 
$$\bar{u}_1 \rightarrow \bar{a}_1 \rightarrow \cdots \rightarrow \bar{u}_n \rightarrow \bar{a}_n$$

We construct an \textit{interaction graph} by merging all such linear paths for all conversations given a specific task. The canonical forms corresponding to the user and agent turns are the \textbf{nodes} of the graph. A directed \textbf{edge} connects each canonical form to the next in the sequence within the conversation. The frequency of a particular transition between two canonical forms (such as $\bar{u}_i \rightarrow \bar{a}_i$ or $\bar{a}_i \rightarrow \bar{u}_{i+1})$, determines the weight of the corresponding edge, i.e. this weight represents the number of occurrences of that transition across all conversations. This results in a weighted directed graph that effectively captures the dynamics of dialogue progression across multiple conversations.

\subsection{Policy Extraction via Graph Traversal}
\label{sec:policy_extraction}
Given the constructed interaction graph, we can extract various \textit{dialogue flows} using graph algorithms. Assuming that a dialogue flow can be represented as a path from a source node to a destination, we can employ various graph traversal algorithms for this stage. Our intuition is that the dialogue policy for the \textit{happy path} of a task should be the most commonly traversed path.
Since the weights of the edges are a proxy for the number of conversations in which that transition occurs, we can consider an algorithm where the objective is to maximize the minimum capacity along a path.

\paragraph{Fattest-Path Dijkstra.} Given a graph depicting a transportation network (graph whose weights are considered as transportation capacities), this is a variation of Dijkstra's algorithm where we want to find a path between the source and the target such that the minimum weight of any edge in the path is as large as possible~\cite{cormen2022introduction}. 

Let \( G = (N, E) \) be a graph with nodes \( N \) and edges \( E \), each edge \( e \) having a weight \( w(e) \). 
Let \( P = \langle s = n_0, n_1, ..., n_k = t \rangle \) be a path from source \( s \) to target \( t \). The bottleneck for path \( P \), denoted by \( F(P) \), is defined as:

\[ F(P) = \min_{0 \leq i < k} \left\{ w(n_i, n_{i+1}) \right\} \]

The goal is to find the path \( P^* \) with the largest bottleneck out of all possible paths from \( s \) to \( t \),  \( \mathcal{P}(s, t) \):

\[ P^* = \argmax_{P \in \mathcal{P}(s, t)} \left\{ F(P) \right\} \]

The source node is chosen as the most commonly occurring canonical form for the first turn observed across all conversations, while the target node is the most frequent final turn canonical form. We then apply the algorithm and extract the fattest-width path as our initial version of the dialogue flow, $df$.

\subsubsection{Dialogue Flow Digressions}

Extracting only the widest path presents us with an incomplete view of the dialogue progression (the \textit{"happy"} or \textit{main path}). We need to find alternative paths arising from nodes on the widest path to capture a more complete dialogue policy.

\begin{figure}[!h]
\centering
\begin{tikzpicture}[level distance=1.5cm,
  level 1/.style={sibling distance=4cm},
  level 2/.style={sibling distance=4cm},scale=0.65]
  \node {\small{Bot ask for account number}}
      child {node {\small{Main path}}}
      child {node [align=right] {\small{User forgot account number}}
        child {node {\small{Initiate recovery process}}}};
\end{tikzpicture}
\caption{Digression from the main path in a graph.}
\label{fig:dialogue-flow-digressions}
\end{figure}
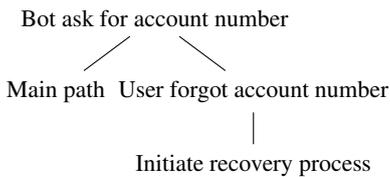

To fully comprehend the necessity of examining digressions, let us consider the example in Fig.~\ref{fig:dialogue-flow-digressions}. Consider a scenario where after the bot asks for the user's account number, the main dialogue path continues with the step \texttt{user provides account number}. However, a potential deviation might occur if the user does not recall the account number. This deviation leads to an alternate path, starting with the canonical form "user forgot account number" and branching into an account recovery subflow. Digressions help us enhance the structure and flow of the dialogue policy.

\paragraph{Identifying Digressions.} For each node in the main dialogue flow, $n_i$, we examine all nodes $n_j$ that are directly connected to $n_{i}$, i.e. $n_{i} \rightarrow n_j$.

To identify potential digressions, we employ a similarity-based thresholding method. If the similarity measure between the canonical form of node $n_j$ and the next node on the main dialogue flow $n_{i+1}$ falls below a specified threshold $\epsilon$, we mark $n_j$ as a digression candidate: 
$$\text{sim}(n_{i+1}, n_j) < \epsilon \implies \text{digression candidate}$$

Then we determine the widest path from each digression candidate node to the final node $n_{end}$ in the dialogue flow. This procedure yields a set of potential alternative paths, denoted as $P_{alt}$. 

Finally, we compute the similarity between each alternative path and the main dialogue trajectory. Paths that exhibit a high similarity to the main dialogue trajectory are discarded, the remaining paths whose similarity to the main flow is below a threshold $\kappa$ are considered digressions. For our experiments, we determined to use $\kappa=0.8$ for selecting digressions by employing a manual evaluation on a small set of extracted digressions with different threshold levels. To compute the similarity between two paths (main and digression) we concatenate the canonical forms between start and end nodes on each path to compute its embedding. 

\section{Experimental Settings}
\label{sec:experiments}
\subsection{Datasets}
We consider two widely used task-oriented dialogue datasets: Schema Guided Dialogue (SGD)~\cite{budzianowski-etal-2018-multiwoz} and Action-Based Conversations Dataset (ABCD)~\cite{chen-etal-2021-action}. 
A relevant aspect for our dialogue policy extraction task is that the conversations were generated using different approaches. SGD dialogues were generated by crowd-sourced paraphrasing of a set of dialogue sketches created automatically driven by a state-machine dialogue policy. Meanwhile, ABCD contains more realistic conversations between a client and a customer support agent (both non-experts) with the agent following a dialogue script resembling real-world customer support scenarios.  

\paragraph{Schema Guided Dialogue (\sgd).} A comprehensive dataset containing 20 domains (or tasks) and 20k annotated conversations. These domains encompass a diverse range of user interactions relevant to an assistant use case, such as setting up calendars, looking for events, and making travel arrangements including different bookings.

\paragraph{Action-Based Conversations Dataset (\abcd).} This dataset is designed to facilitate the development of more realistic customer service dialogue systems, primarily in the e-commerce setting. It contains over 10k human-to-human dialogues, which include the agent taking a specific sequence of actions to accomplish various tasks. The tasks span multiple domains, including managing account details, inquiring about the status of shipping, and handling processes related to initiating and monitoring refunds.

\subsection{Baselines}

Given the limited previous work, we consider the following alternatives for evaluating and comparing the efficacy of the proposed approach (\Scref{sec:method}).

\paragraph{Graph Traversal - Longest Path.} This algorithm identifies the longest path in a directed graph, which, in this context, represents the longest sequence of dialogue turns in a conversation. 
\begin{equation}
L_{max} = \max \left\{ \mathrm{len}(p) : p \in \mathcal{P}(s, t) \right\}
\nonumber
\end{equation}

\paragraph{Graph Traversal - Maximum Weighted Path.}
This method computes the path in a graph that has the highest cumulative weight. In our case, the computed \textit{happy} (main) path would maximize the number of conversations modelled by summing the frequency of each transition on that path. 
\begin{equation}
P_{max} = \max \left\{ \sum_{e \in p} w(e) : p \in \mathcal{P}(s, t) \right\}
\nonumber
\end{equation}

 
\paragraph{Prompting-Based Alternatives.}
Utilizing the set of conversations annotated with canonical forms (Sec. \Scref{sec:intent_identification}) as input, we prompt LLMs to generate the most suitable dialogue policy. We use OpenAI \texttt{gpt-4-turbo} and \texttt{gpt-3.5-turbo}, additional details are in Appendix~\ref{sec:appendix:prompting}. 

\subsection{Evaluation}
\label{sec:evaluation_strategy}
The dialogue policies generated by our method can be expressed in natural language as a sequence of subsequent canonical forms, similar to a conversation. Thus, our automatic evaluation strategy compares a conversation, as a sequence of canonical forms, with the dialogue policy. Additional evaluation details are in Appendix~\ref{sec:appendix:automatic-evaluation} (automatic) and~\ref{sec:appendix:manual-evaluation} (manual, e.g. annotator instructions and interface). 

\subsubsection{Automatic Evaluation}

\paragraph{Text Similarity Metrics.}
Extracting dialogue flows from conversations falls at the intersection of two well-defined language tasks. It can be framed as a translation problem, wherein the goal is to transform unstructured conversational data into a structured dialogue flow format. Additionally, we can view it as a multi-document summarization task~\cite{ma2022multi} involving the distillation of multiple conversations into a concise dialogue flow representation. As the objective is to quantify the ability of the dialogue policy to model the conversations in the corpus, we use standard \textit{text generation metrics} (\bleu, \rouge, \meteor, and \bertscore)~\cite{celikyilmaz2020evaluation} to assess the dialogue flow coverage and quality. To achieve this, we use the canonical forms representation for both conversations and the dialogue policy. 

\paragraph{Structure-Preserving Metric.}
To evaluate how well a dialogue policy respects the structure of conversations and the sequential ordering of canonical forms, we utilize the \textit{Longest Common Subsequence (LCS)}. The longest subsequence common to two sequences can be non-contiguous, but it respects the ordering of elements in each sequence. Let \( C = \{c_1, c_2, \ldots, c_m\} \) represent the sequence of utterances in a conversation, and let \( P = \{p_1, p_2, \ldots, p_n\} \) represent the sequence of actions in an extracted dialogue policy - both encoded as canonical forms. The LCS metric, denoted as \( L(C, P) \), quantifies the number of utterances from conversation \( C \) that can be handled by policy \( P \) in exactly the same order, thereby providing a measure of how well the policy reflects the structure of the conversation.

We compute LCS using two methods: exact match and similarity-based match. Exact match extracts subsequences that have the same canonical forms both in the conversation and the policy, while the similarity-based match uses embedding similarity for matching canonical forms considering two elements a match if their similarity score exceeds a given threshold. This allows a more flexible matching that can correct some of the errors introduced by the intent identification stage.


\subsubsection{Human Evaluation} For an in-depth assessment, we selected five domains from the SGD dataset: the best two for the graph-based method, the best two for the strongest baseline (\texttt{gpt-4-turbo}), and one domain where the performance gap was minimal (see Fig.~\ref{fig:sgd_metrics}). From each domain, we sampled 10 conversations and paired them with the dialogue flows extracted by each method (graph, \texttt{gpt-4-turbo}). Human annotators were then tasked with mapping each step in the dialogue flow to a corresponding turn in the actual conversation - more details are in Appendix~\ref{sec:appendix:manual-evaluation}. In addition to mapping canonical forms in the dialogue flow to the corresponding turn in the conversation, annotators were also asked to rate how relevant the canonical form was to that turn. A score of 1 implied that the canonical form described the user utterance comprehensively, and a score of 0.5 meant that certain details in the utterance were not captured by the canonical form.


\section{Results and Analysis}
\label{sec:analysis}
This section provides a quantitative comparison of graph-based methods and prompting-based techniques. Additionally, a qualitative analysis, examining the variances in extracted dialogue flows, the effect of incorporating digressions into these flows, and the degree of flexibility and control provided by the graph-based approach is also presented.

\begin{table*}[htbp]
\begin{adjustbox}{width=\textwidth, center}
\begin{tabular}{@{}ll|llll|llll@{}}
\toprule
                                 &                        & \multicolumn{4}{c}{SGD}               & \multicolumn{4}{c}{ABCD}           \\ \midrule
                                 &                        & \bleu  & \meteor & \rouge-L & \bertscore & \bleu  & \meteor & \rouge-L & \bertscore \\
\toprule
\multirow{3}{*}{Graph-based (\textbf{\textit{ours}})}     & Longest Path           &  19.16     &  42.94      &    40.76      &  45.04         & 18.39 &  33.12      &  38.45     &  36.46         \\
                                 & Max Weighted Sum       &  19.29     &  42.55      &    40.79      &   44.75        & 19.22 &  35.49      &  39.59     & 37.66          \\
                                 & Fattest-Width Dijkstra & 27.87  & 54.27  & 48.87    & 52.78          & 30.08  & 48.26  & 50.10 &  46.65         \\
                                 & + with 1 Digression & \textbf{28.54}                & \textbf{54.31}                  & \textbf{49.23 }                  & \textbf{52.97 }                    & \textbf{30.93}                & \textbf{48.44}                  & \textbf{50.40}                   & \textbf{47.83}                     \\ \bottomrule
\multirow{2}{*}{Prompting-based} & gpt-3.5-turbo          & 25.52 & 49.77  & 47.54    & 51.37          & 27.64  & 43.54  & 47.88 &  44.31         \\
                                 & gpt-4-turbo            & 26.33  & 52.76  & 48.19    & 52.30          & 28.54  & 45.02  & 48.07 &  44.99        \\ 
\bottomrule
\end{tabular}
\end{adjustbox}
\caption{Comparison of dialogue flow extraction methods using automatic metrics. Fattest path Dijkstra exhibits superior performance over other graph algorithms and surpasses \texttt{gpt-4-turbo} in prompting-based approach across \sgd and \abcd datasets, while adding digressions provides even larger improvement.}
\label{tab:main_results}
\end{table*}

\subsection{Automatic Metrics}
\paragraph{Text Similarity Metrics.}
Table \ref{tab:main_results} presents a comparative analysis of various graph-based and prompting-based methods using text-similarity metrics. In the graph-based category, the Longest-Path and Max-Weighted-Sum methods demonstrate similar performance, with minor variations in scores. The Fattest-width Dijkstra method significantly outperforms both methods in all metrics for both datasets (+8/+12 \bleu, +8/+11 \rouge), indicating a more effective approach in extracting dialogue flows.

\begin{table}[h]
\centering
\small
\begin{tabular}{lccc}
\hline
\textbf{Dataset} & \textbf{\#Turns(Avg)} & \textbf{Graph} & \textbf{\texttt{gpt-4-turbo}} \\
\hline
SGD & 19.10 & 6.10 (7.98) & 5.59 (7.66) \\
ABCD & 11.40 & 4.61 (5.24) & 4.04 (5.00) \\
\hline
\end{tabular}
\caption{Comparison of mean LCS length between extracted (Graph, \texttt{gpt-4-turbo}) policies and conversations. Exact match LCS is outside brackets, similarity-based LCS is inside.}
\label{table:lcs_results}
\end{table}

When compared with prompting-based approaches with LLMs, Fattest-width Dijkstra shows improved performance (+1.5 \bleu, +2 \rouge) over GPT-4 as well. While an optimal prompt might lead to marginally improved scores, the trend suggests that the proposed graph-based method demonstrates a more consistent and effective way to extract dialogue flows.

\paragraph{Structure-Preserving Metric.} Table \ref{table:lcs_results} compares the mean LCS length between policies extracted by the graph and \texttt{gpt-4-turbo} methods. The policies extracted using the graph-based method consistently achieve higher LCS scores, indicating better alignment with the conversations. These results can be interpreted that about a third of all interactions can be correctly handled by the dialogue policies.

\paragraph{Relation to Intent Identification.}
All metrics used for automatic evaluation, including LCS, are dependent on the unsupervised intent identification (\Scref{sec:intent_identification}). Therefore it is important to determine the quality of this step. We have performed a manual analysis and determined that the LLM identifies the user/bot intent with an accuracy of 70\%/87\%. More details can be found in Appendix \Scref{sec:intent_detection_accuracy}. 

\subsection{Human Evaluation}

The annotations from the human evaluation allow us to compute precision and recall metrics to evaluate the extracted dialogue flows. In our context, higher precision indicates that a higher number of canonical forms from the dialogue flow are utilized to describe turns in the conversations. A higher recall implies that a greater number of turns in the conversation are accurately covered by the canonical forms from the dialogue flow. 

\begin{table}[h]
\centering
\begin{tabular}{ccc}
\hline
\textbf{Method} & \textbf{Precision} & \textbf{Recall} \\
\hline
Graph &  73.06 & 65.62 \\
\texttt{gpt-4-turbo} & 68.72 & 64.64 \\
\hline
\end{tabular}
\caption{Precision and Recall between extracted dialogue policies and human annotated conversations.}
\label{table:precision_recall}
\end{table}

Table \ref{table:precision_recall} shows that the dialogue flows extracted using the graph-based approach exhibit significantly higher precision compared to the flows from \texttt{gpt-4-turbo}. This suggests that the graph method is less noisy and more representative of the main flow of the interaction. The recall of the graph-based approach is marginally better than \texttt{gpt-4-turbo} indicating similar efficacy in capturing conversation turns. Table \ref{table:performance} in Appendix \Scref{sec:precision_recall_policies} shows a breakdown of performance in terms of precision and recall of the user and bot canonical forms.

\begin{table}[h]
\centering
\begin{tabular}{lcc}
\hline
\textbf{Domain}         & \textbf{Graph} & \textbf{\texttt{gpt-4-turbo}} \\
\hline
BookAppointment & 0.59 & 0.51 \\
SearchHotel & 0.60 & 0.53 \\
ReserveRestaurant & 0.55 & 0.57 \\
GetEventDates & 0.43 & 0.52 \\
PlayMedia & 0.37 & 0.49 \\
\hline
\end{tabular}
\caption{Comparison of Graph and \texttt{gpt-4-turbo} scores for the 5 domains in SGD used for human evaluation.}
\label{tab:avg_human_scores}
\end{table}


The average scores for each domain are shown in Table \ref{tab:avg_human_scores}. We observe that the trends of the manually annotated scores are consistent with automatic evaluations (see Fig.~\ref{fig:sgd_metrics}). Domains where the graph method outperforms \texttt{gpt-4-turbo} in the automatic evaluation, such as \textit{'BookAppointment'} and \textit{'SearchHotel'}, are reflected similarly by human annotators. This indicates a strong correlation between automatic metrics and human ratings. The lower scores for the graph method in specific areas can be attributed to the lack of digressions in the main dialogue flows used for human evaluation. Future improvements addressing this aspect could enhance the efficacy of the graph method.

\subsection{Considerations for Developers}

The proposed graph-based methodology offers several advantages over prompting-based techniques, particularly in terms of control and flexibility for conversation designers.

\paragraph{Controllability.} Graph-based methods provide superior control, allowing designers to specify dialogue flow length, identify digressions, and decide which digressions to include. Prompt-based methods lack this fine-tuned control and interpretability, making precise modifications challenging. Graph-based methods allow control over dialogue flow length, allowing developers to balance precision and recall effectively. 

\paragraph{Adding Digressions.} Integrating digressions into dialogue flows enhances understanding of conversational dynamics. As shown in Table~\ref{tab:main_results}, adding a single digression improves all metrics by about 1 point. Graph-based methods facilitate precise identification and mapping of digressions, offering a clear visual representation of dialogue progression, which is beneficial for conversation modelling.

\paragraph{Robustness.} Prompting-based approaches can be brittle and influenced by the order of conversation presentation, leading to inconsistent results. Graph-based methods produce deterministic outputs, ensuring predictable and consistent results regardless of input order, which is crucial for reliable conversation design.

\section{Conclusion}
\label{sec:conclusion}
Generating dialogue policies from a dataset of conversations can significantly reduce the effort required by conversation designers and domain experts to develop TODS. We propose a novel hybrid LLM and graph-based method to extract dialogue policies without relying on a predefined set of dialogue acts.

Our results are significant for three reasons. First, we demonstrate that dialogue policies can be computed using network flow in a graph of all possible conversations for a given task. Second, modeling conversations as sequences of canonical forms enhances explainability and controllability. Third, incorporating digressions as high-flow paths in the graph allows conversation designers to control the granularity of dialogue policies.
\section{Limitations and Risks}
\label{sec:limitations}
The dialogue policies generated with the proposed approach are not perfect and should not be used to implement any TODS without careful inspection by a conversation designer or domain expert. Moreover, we acknowledge that in most cases the extracted dialogue flows will be iteratively improved by human experts. Therefore, our method is mainly intended to serve as a productivity tool. As the generated policies are expressed as sequences of canonical forms expressed in English it provides a good degree of explainability for the generated dialogue policies.  At the same time, the mechanism for identifying digressions helps control the granularity and coverage of the dialogue policies and can be used by experts to analyze existing datasets. 

A further limitation of our research is that we have not fully investigated the impact of various intent identification methods. For example, the clustering algorithm and sentence embeddings used by the intent normalization stage might influence the performance of our graph-based method. At the same time, other intent extraction methods described in Section \Scref{sec:background} should also be compared to our proposed method. All these will go into future work and experiments. 

At last, while in our work we have shown that the automatic evaluation using text generation metrics (e.g. \bleu, \bertscore) are correlated very well with the human evaluation on 5 different conversational tasks, this may not be the case on other conversation datasets. Therefore, we encourage developers that want to use this approach for evaluating the performance of the extracted dialogue flows to check first that the automatic metrics are well correlated with (at least a small) human annotated dataset that measures overlap between policies and conversations with domain experts.

The main risks of our approach is that the generated dialogue policies might contain canonical forms that are irrelevant or even malicious, but are extracted somehow from the corpus of conversations offered as input. However, we consider this should not be the case as the extracted policies should always be investigated and curated by a conversation designer.

\section{Broader Impact}
Upon acceptance for publication, we aim to release both the code and the generated dialogue policies for the ABCD and SGD datasets. In accordance with OpenAI terms of usage, this data would be available only for research purposes and would not be commercially usable. We also aim to release a set of conversations that are annotated with Llama3-70B-Instruct to enable better annotation quality. We consider that the existence of such a tool for extracting dialogue policies would benefit companies and developers that have access to datasets of task-oriented conversations. 

\bibliography{anthology,custom}

\clearpage
\appendix
\section{Implementation Details for Intent Identification}
\label{sec:appendix:intent_identification}

\subsection{Prompt Template for Creating Weak Labels.} As mentioned in \Scref{sec:intent_identification}, we have used \texttt{text-davinci-003} from OpenAI to create an initial set of weak labels for 200 conversations from the ABCD dataset on 2 different tasks. For this, we have used the following prompt:

\begin{Verbatim}[breaklines=true,breaksymbolleft=]
Your task is to annotate conversational utterances with intents expressed as canonical forms. Canonical forms are short summaries representing the intent of the utterance - it is neither too verbose nor too short. Here is an example to show you how the task is to be performed.

{example}

Annotate the following conversation in a similar manner. if similar intents are detected, make sure to use the same canonical forms as in the example given. for other ones, use the ones in the example above as reference and craft them. Each turn of the conversation should be annotated with the corresponding canonical forms.

{conv}

Output the annotated conversation with canonical forms.
\end{Verbatim}


The usage of \texttt{text-davinci-003} was based on its performance in generating canonical forms used for dialogue rails in NeMo Guardrails~\cite{rebedea2023nemo}. While the model has been deprecated at the end of 2023, initial experiments show that the new model, \texttt{gpt-3.5-turbo-instruct} achieves a similar performance for this task. For all runs, we have used greedy decoding with temperature equal to 0. 

\subsection{P-Tuned LLM for Intent Identification with Weak Labels}
\label{sec:ptuned_llm}
For this study, we make use of our in-house 43 billion parameter model as the base LM. The 43B model is a decoder-only GPT architecture LLM that has been trained on 1.1 trillion tokens. It has 48 layers and uses a vocabulary size of 256 thousand, RoPE positional embeddings~\cite{lee2021rope} and SwiGLU activation~\cite{shazeer2020glu} without dropout. It was aligned using a combination of publicly available and proprietary alignment data. For p-tuning, we used a batch size of 8, learning rate of \texttt{1e-4}, number of virtual tokens as 30 and trained for 50 epochs with early stopping. The best performance was obtained at epoch 20. The training data is structured such that the model is trained to predict the canonical form for a particular turn given the conversation history up to that point. The training data consists of 850 samples and the validation data consists of 300 samples.

\subsection{Intent Detection Performance of the p-tuned LLM}
\label{sec:intent_detection_accuracy}
To test the accuracy of intent identification, we manually annotate with canonical forms the turns in 20 conversations. Then we compare these annotations with the predictions from the p-tuned LLM for canonical forms extraction. We use MiniLM-v6 to measure the semantic similarity between the predicted canonical form by the p-tuned LLM and the ground truth. If this similarity score exceeds 0.8, the prediction is marked as correct; otherwise, it is classified as incorrect. Based on this method, we calculate the user intent accuracy at 70\% and the bot intent accuracy at 87\%. When employing \texttt{text-davinci-003} for generating the predicted canonical forms, the user intent accuracy improves to 86\%, and the bot intent accuracy to 93\%.

Exploring the usage of more annotated data for p-tuning and the use of better LLMs like \texttt{Llama-3-70B-Instruct} would lead to significantly better intent identification results. 

\subsection{Intent Normalization}
For SGD, we use agglomerative clustering with a clustering threshold of 0.9 and Euclidean distance as the metric. Similarly, for ABCD dataset the clustering threshold is set at 0.7. We select these clustering thresholds after running a hyperparameter search over a range of clustering threshold values (0.5-1.0).

\section{Examples of Generated Dialogue Policies}
\label{sec:appendix:examples-flows}

In Table~\ref{tab:sample-dialogue-flows} we show the dialogue policies extracted with the proposed graph-based method for 4 domains from the SGD dataset. For each domain, we can compare the main (happy) path with a flow containing one additional digression added to the main path. To tackle the branching of a flow (e.g. digression vs. main path) we are using some simple syntactic features supported by Colang flows~\cite{rebedea2023nemo}, i.e. the special keyword \texttt{when}. This works by traversing the digression path only when the specific user intent in the \texttt{when} condition is met, otherwise continuing with the main path.

In addition to the dialogue flows, each domain also has a sample conversation (out of several hundreds) used to generate the flows. 


\section{Details for Prompting-Based Dialogue Policy Generation}
\label{sec:appendix:prompting}

The following methodology has been used to generate the dialogue policy using prompting given the corpus of conversations for a task. 

After several iterations, we have used the following prompt which provides good results for the task of generating a dialogue flow from a set of conversations modelled using canonical forms.

\begin{Verbatim}[breaklines=true,breaksymbolleft=]
Here is a list of dialogue flows that denote how conversations usually proceed between a user and a bot. Your task is to create a dialogue flow that best represents the conversation flow given all the dialogue flows below.

{conversations_with_canonical_forms}

What is the most commonly traversed path in this set of conversations? Output it following a similar format as the conversations above. Only display the output path. Do not add any comments or other text.
\end{Verbatim}


Due to context length limitations, we utilize a batch of 100 conversations as input for the LLM (and the graph method).  Following this, we extract the dialogue flow from these conversations.

\section{Automatic Evaluation Metrics by Domain}
\label{sec:appendix:automatic-evaluation}
The evaluation is always conducted using a distinct set of conversations that were not included in the batch of 100 conversations used for extracting the dialogue flows. For example, in a domain with 300 conversations, the dialogue flow is extracted from the 100 conversations at a time and evaluated against the remaining 200. This helps ensure a fair evaluation and mitigates the risk of overfitting.

\begin{figure*}
    \centering
    \begin{subfigure}{0.45\textwidth}
        \centering
        \includegraphics[width=\linewidth]{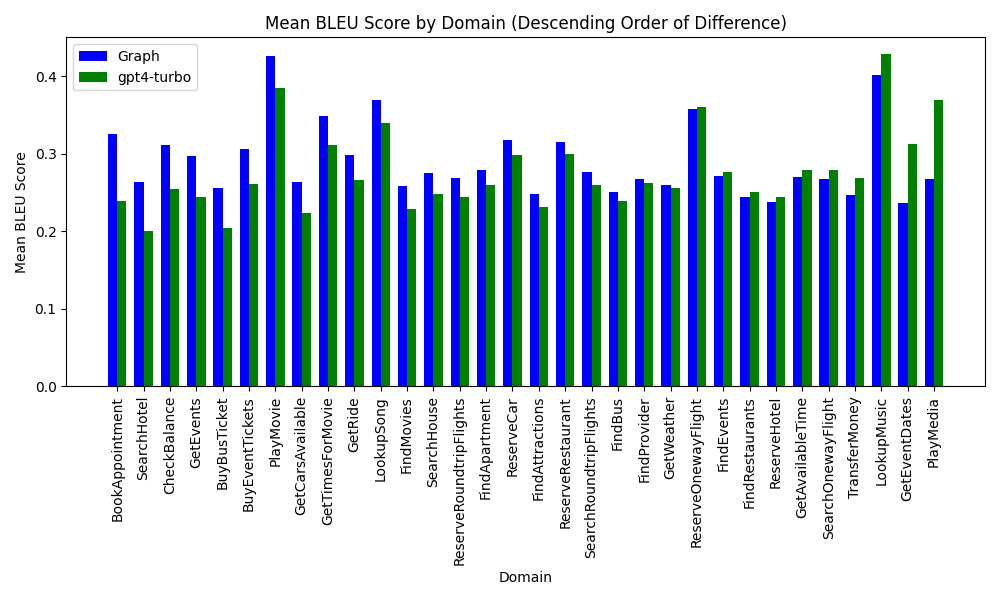}
        \caption{BLEU}
        \label{fig:sub1}
    \end{subfigure}
    \hfill
    \begin{subfigure}{0.45\textwidth}
        \centering
        \includegraphics[width=\linewidth]{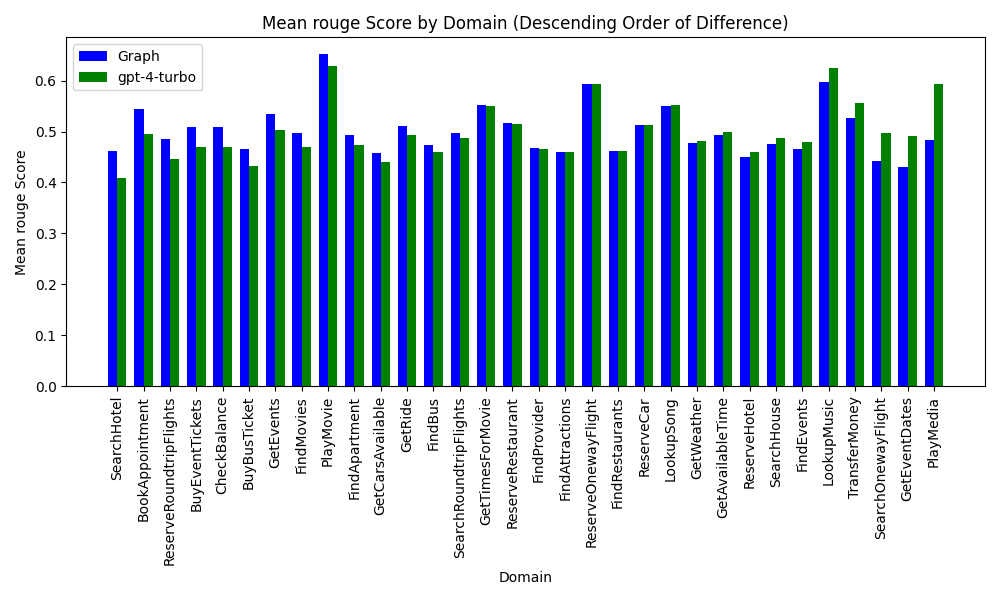}
        \caption{ROUGE-L}
        \label{fig:sub2}
    \end{subfigure}
    \caption{Performance obtained by the proposed graph-based method and the top performing baseline (\texttt{gpt-4-turbo}) for the SGD dataset. The results are ordered from left to right on the value of the difference in performance between graph-based and GPT-4.}
    \label{fig:sgd_metrics}
\end{figure*}

\begin{figure*}
    \centering
    \begin{subfigure}{0.45\textwidth}
        \centering
        \includegraphics[width=\linewidth]{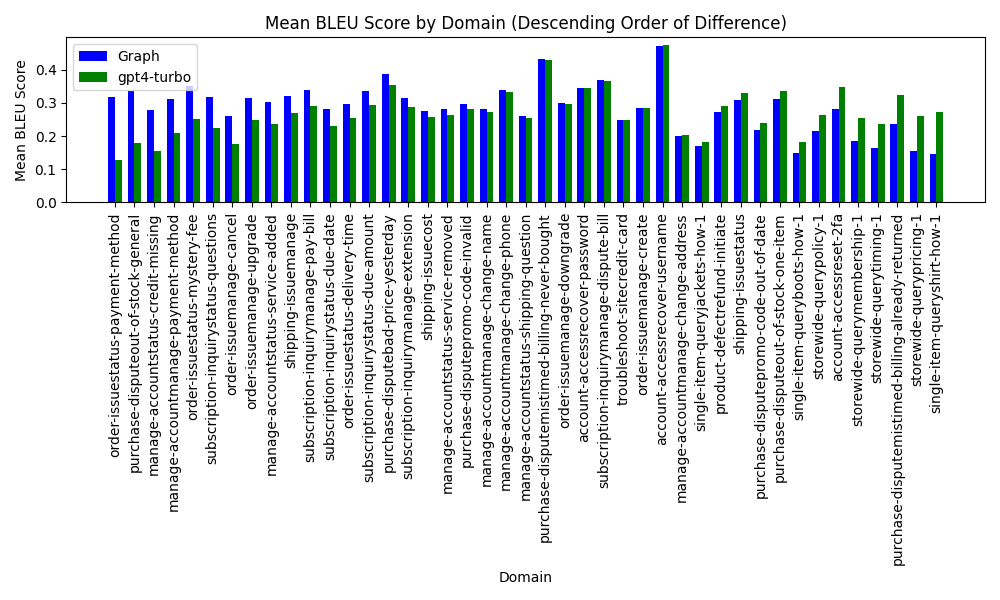}
        \caption{BLEU}
        \label{fig:abcd_sub1}
    \end{subfigure}
    \hfill
    \begin{subfigure}{0.45\textwidth}
        \centering
        \includegraphics[width=\linewidth]{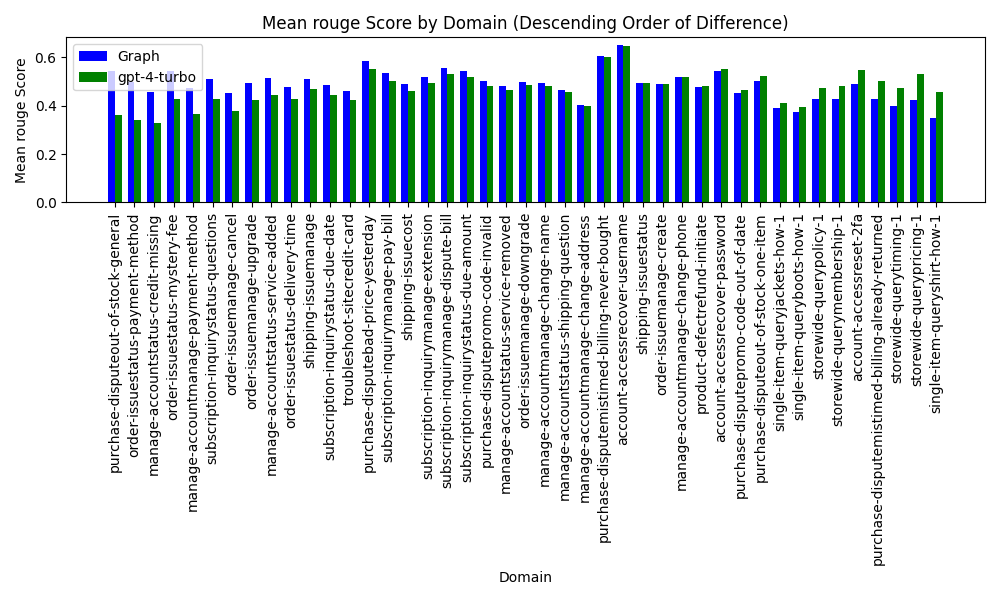}
        \caption{ROUGE-L}
        \label{fig:abcd_sub2}
    \end{subfigure}
    \caption{Performance obtained by the proposed graph-based method and the top performing baseline (\texttt{gpt-4-turbo}) for the ABCD dataset. The results are ordered from left to right on the value of the difference in performance between graph-based and GPT-4.}
    \label{fig:abcd_metrics}
\end{figure*}

In Figs.~\ref{fig:sgd_metrics} and~\ref{fig:abcd_metrics} we provide the \bleu and \rouge scores for each task in the SGD and ABCD datasets for the top two performing methods: Graph and GPT-4. In each graph, the tasks are ordered from left to right based on the value of the difference in performance on that metric between the proposed graph-based method and GPT-4 prompting. We can easily see that the Graph method is out-performing GPT-4 in more than 60\% of the tasks for both datasets.

\section{Manual Evaluation}
\label{sec:appendix:manual-evaluation}

\begin{figure*}
    \centering
    \includegraphics[width=\textwidth]{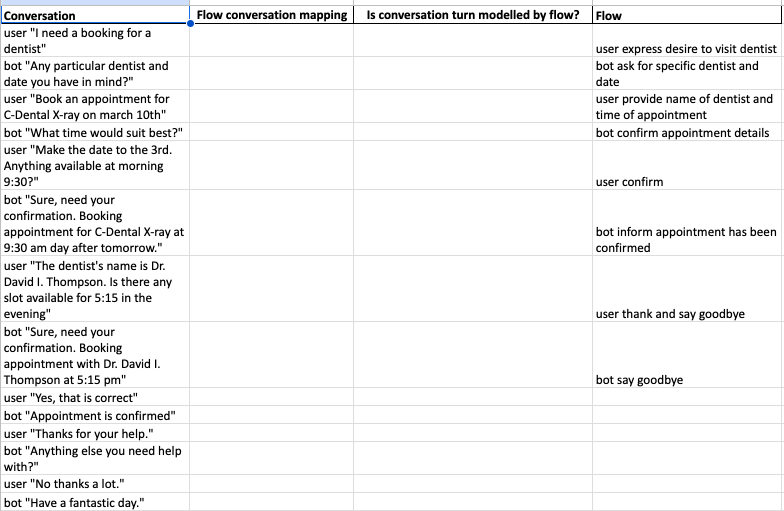}
    \caption{Annotation UI for the Manual Evaluation}
    \label{fig:annotation-ui}
\end{figure*}

For manual evaluation, we use 8 annotators. For this task, we selected volunteers instead of relying on crowd-workers. Each volunteer has at least a MSc in Computer Science or related domain, being at least knowledgeable in NLP. Each annotator was tasked with reviewing 25 pairs of conversation and associated dialogue flow. More, annotators received a balanced distribution between flows generated by the Graph-based method and by GPT-4. The annotators are asked to map the canonical forms in the dialogue flow to the corresponding conversation turn, as well as assign a score to the canonical form (\ref{sec:evaluation_strategy}). The annotation UI is shown in Fig.~\ref{fig:annotation-ui}. The annotators are not made aware of which method is used to extract the dialogue flow to prevent any potential bias. 

Before starting the annotations, one of the authors of the paper provided about 10 pairs of conversation and dialogue flow as samples annotations and also a short guide of about 3-4 pages on the annotation process. The guide included an explanation of the task, the annotation UI, and had a short lost of Q\&A. The time required per annotator was about 2 hours and the annotators were paid for this task.

In order to have consistent manual annotations, each pair of conversation and dialogue flow was labelled by two different annotators. We have obtained a substantial inter-rater agreement, Cohen's $\kappa$=0.71, considering a binary classification task for the canonical forms in the dialogue flow (matched or not matched by a turn in the current conversation). 

\subsection{Precision and Recall of Policies}
\label{sec:precision_recall_policies}
Let us assume that we have a sample conversation that goes like
\begin{itemize}
    \item \textbf{User}: "What's the weather like today?"\\
\textbf{Bot}: "The weather is sunny with a high of 75 degrees."\\
\textbf{User}: "Will it rain tomorrow?"\\
\textbf{Bot}: "No, it is expected to be clear all day tomorrow."\\
\textbf{User}: "What about this weekend?"\\
\textbf{Bot}: "It might rain on Saturday, but Sunday should be sunny."
\end{itemize}

Predicted dialogue trajectory for this conversation is
\begin{verbatim}
user ask about weather
bot provide weather
user ask about weather tomorrow
bot provide weather
user ask about weather weekend
bot provide weather
bot provide weather
\end{verbatim}

The human annotator is tasked with mapping turns in the conversation with the appropriate intent/canonical form from the dialogue trajectory.
\begin{itemize}
    \item \textbf{User}: "What's the weather like today?"
    $\rightarrow$ \textit{user ask about weather}
    
    \item \textbf{Bot}: "The weather is sunny with a high of 75 degrees."
    $\rightarrow$ \textit{bot provide weather}
    
    \item \textbf{User}: "Will it rain tomorrow?"
    $\rightarrow $\textit{user ask about weather tomorrow}
    
    \item \textbf {Bot}: "No, it is expected to be clear all day tomorrow."
    $\rightarrow$ \textit{bot provide weather}
    
    \item \textbf{User}: "What about this weekend?" \\
    $\rightarrow$ \textit{user ask about weather weekend}
    
    \item \textbf{Bot}: "It might rain on Saturday, but Sunday should be sunny."
    $\rightarrow$ \textit{bot provide weather}
\end{itemize}

Once we have this mapping, we see that the user canonical forms encode the state (similar to intent and slots in a standard NLU), while the bot canonical forms measure how well the predicted responses from the extracted policy match the bot responses in the conversation under evaluation. We then evaluate the performance in terms of micro-precision and micro-recall of the graph-based and prompt-based approaches on the "user" canonical forms and the "bot" canonical forms of the extracted policies. Table \ref{table:performance} shows the precision and recall breakdown for both the graph and prompt-based approaches. We see that the graph-based approach obtains a higher precision and recall score compared to the prompt-based method.

\begin{table}[!htbp]
\centering
\begin{adjustbox}{width=\linewidth, center}
\begin{tabular}{@{}lllll@{}}
\toprule
\textbf{Metric} & \textbf{Minimum} & \textbf{Maximum} & \textbf{Average} & \textbf{Standard Deviation} \\ 
\midrule
\bleu            & 25.6             & 26.5             & 26.1             & 0.3                         \\
\rouge-L           & 47.7             & 48.4             & 47.9             & 0.26                         \\
\bottomrule
\end{tabular}
\end{adjustbox}
\caption{Variance in \bleu and \rouge-L metrics across 5 runs of the prompting-based method using \texttt{gpt-4-turbo}.}
\label{tab:metrics}
\end{table}

\begin{table*}[htbp]
\small
\centering
\begin{tabular}{|l|c|c|c|c|}
\hline
\textbf{} & \textbf{User - Precision} & \textbf{User - Recall} & \textbf{Bot - Precision} & \textbf{Bot - Recall} \\
\hline
\textbf{Graph} & 76.5 & 68.3 & 78.8 & 68.4  \\
\hline
\textbf{gpt-4-turbo} & 71.4 & 66.8 & 73.3 & 67.5 \\
\hline
\end{tabular}
\caption{Performance metrics of the graph-based approach and gpt-4-turbo model in terms of precision and recall for user and bot canonical forms when compared with human annotations.}
\label{table:performance}
\end{table*}

\section{Variance in LLM output}
\label{sec:appendix:variance_llm}
To evaluate the effect of altering the sequence of conversations, we prompted \texttt{gpt-4-turbo} to extract the dialogue flows for all domains in the \sgd dataset. This process was repeated five times for each domain, with the same set of conversations in the prompt, but with their order randomized in each iteration.

Table \ref{tab:metrics} indicates a relatively narrow range in both \bleu and \rouge scores indicating consistent performance. However, determinism is preferred as it ensures reproducibility and reliability in the results.

\begin{table*}[htbp]
\begin{adjustbox}{width=\textwidth, center}
\begin{tabular}{@{}llll@{}}
\toprule
Domain     & Dialogue Flow                                                                                                                                                                                                                                                                                                                                                                                                                                                                      & Dialogue Flow with Digression                                                                                                                                                                                                                                                                                                                                                                                                                                                                                                                                                                                   & Conversation                                                                                                                                                                                                                                                                                                                                                                                                                                                                                                                                                                                                                                                                                                                                                                                                                                                                                                                                                                                                                                                                                                                                                                                                                 \\ \midrule
PlayMedia  & \begin{tabular}[c]{@{}l@{}}\textcolor{blue}{user} request to play song\\ \textcolor{red}{bot} ask for song to play\\ \textcolor{blue}{user} provide song name\\ \textcolor{red}{bot} confirm request to play song\\ \textcolor{blue}{user} confirm\\ \textcolor{red}{bot} inform movie is playing\\ \textcolor{blue}{user} thank\\ \textcolor{red}{bot} ask if any other help is needed\\ \textcolor{blue}{user} thank and say goodbye\\ \textcolor{red}{bot} say goodbye\end{tabular}                                                                                                                                                                            & \begin{tabular}[c]{@{}l@{}}\textcolor{blue}{user} request to play song\\ \textcolor{red}{bot} ask for song to play\\ \textcolor{blue}{user} provide song name\\ \textcolor{red}{bot} confirm request to play song\\ \textcolor{blue}{user} confirm\\ \textcolor{red}{bot} inform movie is playing\\ \textit{\textbf{when}} \textcolor{blue}{user} request to make transfer\\ \textcolor{red}{bot} ask for amount to transfer\\ \textcolor{blue}{user} provide amount to transfer\\ \textcolor{red}{bot} confirm transfer request\\ \textcolor{blue}{user} confirm\\ \textcolor{red}{bot} inform movie is playing\\ \textcolor{blue}{user} thank\\ \textcolor{red}{bot} ask if any other help is needed\\ \textcolor{blue}{user} thank and say goodbye\\ \textcolor{red}{bot} say goodbye\end{tabular}                                                                                                                     & \begin{tabular}[c]{@{}l@{}}\textcolor{blue}{user} "I wish to play the song. Broadcast it on my display. I love Kacey Musgraves song."\\ \textcolor{red}{bot} "Which song do you need?"\\ \textcolor{blue}{user} "I need the song Slow Burn."\\ \textcolor{red}{bot} "Kindly ensure that, Slow Burn is will play on TV."\\ \textcolor{blue}{user} "Yes. it is perfect."\\ \textcolor{red}{bot} "Song is started."\\ \textcolor{blue}{user} "Thank you so much."\\ \textcolor{red}{bot} "Do you need any other help?"\\ \textcolor{blue}{user} "I wish to make a transfer amount from my checking account to savings account of Jerry."\\ \textcolor{red}{bot} "How much to be transferred?"\\ \textcolor{blue}{user} "Please transfer 1110 bucks."\\ \textcolor{red}{bot} "Kindly ensure that, \$1,110 is transferred from my checking account to savings account of Jerry."\\ \textcolor{blue}{user} "Yes. It is correct."\\ \textcolor{red}{bot} "The transfer is made successfully."\\ \textcolor{blue}{user} "I wish to play the movie on internet now."\\ \textcolor{red}{bot} "What movie do you wish to watch?"\\ \textcolor{blue}{user} "Play Mary Magdalene with subtitles."\\ \textcolor{red}{bot} "Kindly ensure that, playing Mary Magdalene without subtitles."\\ \textcolor{blue}{user} "Yes. What type of genre it is? Who is the director?"\\ \textcolor{red}{bot} "Movie started. It is a Drama movie directed by Garth Davis."\\ \textcolor{blue}{user} "Thank you so much."\\ \textcolor{red}{bot} "Do you need any other help?"\\ \textcolor{blue}{user} "No. Thank you so much."\\ \textcolor{red}{bot} "Have a fantastic day."\end{tabular} \\ \bottomrule
GetWeather & \begin{tabular}[c]{@{}l@{}}\textcolor{blue}{user} request weather information\\ \textcolor{red}{bot} ask for city\\ \textcolor{blue}{user} provide city\\ \textcolor{red}{bot} provide weather forecast\\ \textcolor{blue}{user} request music\\ \textcolor{red}{bot} recommend song\\ \textcolor{blue}{user} acknowledge recommend song\\ \textcolor{red}{bot} ask if song should be played\\ \textcolor{blue}{user} confirm request to play song\\ \textcolor{red}{bot} confirm request to play song\\ \textcolor{blue}{user} confirm\\ \textcolor{red}{bot} inform song is playing\\ \textcolor{blue}{user} thank\\ \textcolor{red}{bot} ask if any other help is needed\\ \textcolor{blue}{user} thank and say goodbye\\ \textcolor{red}{bot} say goodbye\end{tabular} & {\color[HTML]{330001} \begin{tabular}[c]{@{}l@{}}\textcolor{blue}{user} request weather information\\ \textcolor{red}{bot} ask for city\\ \textcolor{blue}{user} provide city\\ \textcolor{red}{bot} provide weather forecast\\ \textit{\textbf{when}} \textcolor{blue}{user} ask for humidity\\ \textcolor{red}{bot} provide humidity\\ \textcolor{blue}{user} request music\\ \textcolor{red}{bot} recommend song\\ \textcolor{blue}{user} acknowledge recommend song\\ \textcolor{red}{bot} ask if song should be played\\ \textcolor{blue}{user} confirm request to play song\\ \textcolor{red}{bot} inform song is playing\\ \textcolor{blue}{user} thank\\ \textcolor{red}{bot} ask if any other help is needed\\ \textcolor{blue}{user} thank and say goodbye\\ \textcolor{red}{bot} say goodbye\end{tabular}}                                                                                                  & \begin{tabular}[c]{@{}l@{}}\textcolor{blue}{user} "I need the weather forecast for the 7th."\\ \textcolor{red}{bot} "What city would you like to check the weather?"\\ \textcolor{blue}{user} "Please check Ciudad de Mexico"\\ \textcolor{red}{bot} "For that day there is a 6 percent chaning of rain and the average temperature should be 93 degrees Fahrenheit."\\ \textcolor{blue}{user} "What will the humidity be? Will it be windy?"\\ \textcolor{red}{bot} "The humidity will be around 24 percent and the average wind speed 5 miles per hour."\\ \textcolor{blue}{user} "Thanks. Can you sind some songs for me?"\\ \textcolor{red}{bot} "I found 10 songs. Do you like Stitches by Shawn Mendes on the album Handwritten?"\\ \textcolor{blue}{user} "That sounds great. Please play the song."\\ \textcolor{red}{bot} "Please confirm: play Stitches on the TV."\\ \textcolor{blue}{user} "That is correct. What genre is that?"\\ \textcolor{red}{bot} "The song is now playing. It's a pop song."\\ \textcolor{blue}{user} "Thanks."\\ \textcolor{red}{bot} "Can I assist you with something else?"\\ \textcolor{blue}{user} "No thanks."\\ \textcolor{red}{bot} "Enjoy your day."\end{tabular}                                                                                                                                                                                                                                                                                                       \\ \bottomrule
LookupSong & \begin{tabular}[c]{@{}l@{}}\textcolor{blue}{user} request to find a song\\ \textcolor{red}{bot} suggest song\\ \textcolor{blue}{user} confirm song recommendation\\ \textcolor{red}{bot} ask if \textcolor{blue}{user} wants to play song\\ \textcolor{blue}{user} request to play song on kitchen speaker\\ \textcolor{red}{bot} confirm request to play song on kitchen speaker\\ \textcolor{blue}{user} confirm\\ \textcolor{red}{bot} inform song is playing\\ \textcolor{blue}{user} thank\\ \textcolor{red}{bot} ask if any other help is needed\\ \textcolor{blue}{user} thank and say goodbye\\ \textcolor{red}{bot} say goodbye\end{tabular}                                                                  & \begin{tabular}[c]{@{}l@{}}\textcolor{blue}{user} request to find a song\\ \textcolor{red}{bot} suggest song\\ \textit{\textbf{when}} \textcolor{blue}{user} ask for genre\\ \textcolor{red}{bot} provide genre\\ \textcolor{blue}{user} confirm song recommendation\\ \textcolor{red}{bot} ask if \textcolor{blue}{user} wants to play song\\ \textcolor{blue}{user} request to play song on kitchen speaker\\ \textcolor{red}{bot} confirm request to play song on kitchen speaker\\ \textcolor{blue}{user} confirm\\ \textcolor{red}{bot} inform song is playing\\ \textcolor{blue}{user} thank\\ \textcolor{red}{bot} ask if any other help is needed\\ \textcolor{blue}{user} thank and say goodbye\\ \textcolor{red}{bot} say goodbye\end{tabular}                                                                                                                                                 & \begin{tabular}[c]{@{}l@{}}\textcolor{blue}{user} "I want to hear a song from nine years back."\\ \textcolor{red}{bot} "How about Alone by Alan Walker from the album Different World?"\\ \textcolor{blue}{user} "What genre of music is that?"\\ \textcolor{red}{bot} "It's labeled as a Pop song."\\ \textcolor{blue}{user} "Okay, that one will work. Please play it on the Kitchen speaker."\\ \textcolor{red}{bot} "To confirm: You want to play Alone on the Kitchen speaker."\\ \textcolor{blue}{user} "Sorry, no. Please play it on the speaker in the bedroom."\\ \textcolor{red}{bot} "Okay. So, I'll play the song Alone on the Bedroom speaker."\\ \textcolor{blue}{user} "That's right. Please go ahead."\\ \textcolor{red}{bot} "Sure. Enjoy the music!"\\ \textcolor{blue}{user} "Thanks. That's all for now."\\ \textcolor{red}{bot} "Have a wonderful day!"\end{tabular}                                                                                                                                                                                                                                                                                                                                                                                                                                                                                                                                                      \\ \bottomrule
FindMovies & \begin{tabular}[c]{@{}l@{}}\textcolor{blue}{user} express desire to watch a movie\\ \textcolor{red}{bot} ask for location\\ \textcolor{blue}{user} provide location\\ \textcolor{red}{bot} suggest movies\\ \textcolor{blue}{user} confirm\\ \textcolor{red}{bot} ask if \textcolor{blue}{user} wants to buy tickets\\ \textcolor{blue}{user} decline to buy tickets\\ \textcolor{red}{bot} ask if any other help is needed\\ \textcolor{blue}{user} thank and say goodbye\\ \textcolor{red}{bot} say goodbye\end{tabular}                                                                                                                                                          & \begin{tabular}[c]{@{}l@{}}\textcolor{blue}{user} express desire to watch a movie\\ \textcolor{red}{bot} ask for location\\ \textcolor{blue}{user} provide location\\ \textcolor{red}{bot} suggest movies\\ \textit{\textbf{when}} \textcolor{blue}{user} ask for showtime\\ \textcolor{red}{bot} provide showtime\\ \textcolor{blue}{user} acknowledge provide showtime\\ \textcolor{blue}{user} confirm\\ \textcolor{red}{bot} ask if \textcolor{blue}{user} wants to buy tickets\\ \textcolor{blue}{user} decline to buy tickets\\ \textcolor{red}{bot} ask if any other help is needed\\ \textcolor{blue}{user} thank and say goodbye\\ \textcolor{red}{bot} say goodbye\end{tabular}                                                                                                                                                                                               & \begin{tabular}[c]{@{}l@{}}\textcolor{blue}{user} "Look for a movie for me to watch."\\ \textcolor{red}{bot} "Where are you located?"\\ \textcolor{blue}{user} "Find me movies in Union City."\\ \textcolor{red}{bot} "There are 10 movies there. What do you think about Breakthrough, Captain Marvel or After?"\\ \textcolor{blue}{user} "Captain Marvel sounds great. At what time could I watch it?"\\ \textcolor{red}{bot} "What date do you want to see it on?"\\ \textcolor{blue}{user} "I'd like to watch it on the 7th of March."\\ \textcolor{red}{bot} "I found 1 showtime for Captain Marvel. There is one at 9 pm in Century 25 Union Landing \& XD."\\ \textcolor{blue}{user} "Sounds good."\\ \textcolor{red}{bot} "Do you want to buy tickets?"\\ \textcolor{blue}{user} "No, I do not want to buy them right now."\\ \textcolor{red}{bot} "Do you need any more help?"\\ \textcolor{blue}{user} "No, thanks."\\ \textcolor{red}{bot} "Enjoy your day."\end{tabular}                                                                                                                                                                                                                                                                                                                                                                                                                                                                                              \\ \bottomrule
GetRide    & \begin{tabular}[c]{@{}l@{}}\textcolor{blue}{user} request for cab\\ \textcolor{red}{bot} offer shared ride\\ \textcolor{blue}{user} provide number of passengers and request shared ride\\ \textcolor{red}{bot} ask for destination\\ \textcolor{blue}{user} provide destination\\ \textcolor{red}{bot} ask for shared ride\\ \textcolor{blue}{user} confirm shared ride\\ \textcolor{red}{bot} confirm request for shared ride\\ \textcolor{blue}{user} confirm\\ \textcolor{red}{bot} inform ride has been booked\\ \textcolor{blue}{user} thank\\ \textcolor{red}{bot} ask if any other help is needed\\ \textcolor{blue}{user} thank and say goodbye\\ \textcolor{red}{bot} say goodbye\end{tabular}                                & \begin{tabular}[c]{@{}l@{}}\textcolor{blue}{user} request for cab\\ \textit{\textbf{when}} \textcolor{red}{bot} ask for number of seats\\ \textcolor{blue}{user} provide number of seats\\ \textcolor{red}{bot} ask for ride type\\ \textcolor{blue}{user} provide preferred ride type\\ \textcolor{red}{bot} confirm request for ride\\ \textcolor{red}{bot} offer shared ride\\ \textcolor{blue}{user} provide number of passengers and request shared ride\\ \textcolor{red}{bot} ask for destination\\ \textcolor{blue}{user} provide destination\\ \textcolor{red}{bot} ask for shared ride\\ \textcolor{blue}{user} confirm shared ride\\ \textcolor{red}{bot} confirm request for shared ride\\ \textcolor{blue}{user} confirm\\ \textcolor{red}{bot} inform ride has been booked\\ \textcolor{blue}{user} thank\\ \textcolor{red}{bot} ask if any other help is needed\\ \textcolor{blue}{user} thank and say goodbye\\ \textcolor{red}{bot} say goodbye\end{tabular} & \begin{tabular}[c]{@{}l@{}}\textcolor{blue}{user} "I need to go to 19369 Stevens Creek Boulevard Suite 130 for which I need a cab with Luxury ride. Can you find one?"\\ \textcolor{red}{bot} "Can you tell me the number of seats that you need?"\\ \textcolor{blue}{user} "Yes, I need only one."\\ \textcolor{red}{bot} "Please confirm to book a Luxury cab with 1 seat to 19369 Stevens Creek Boulevard Suite 130."\\ \textcolor{blue}{user} "can you check for a Cab with pool ride for 4 persons?"\\ \textcolor{red}{bot} "please confirm to book a Cab with Pool ride with 4 seats."\\ \textcolor{blue}{user} "Yes, that is okay for me."\\ \textcolor{red}{bot} "Your booking was confirmed and the Cab is on the way to your location."\\ \textcolor{blue}{user} "Can you tell me the exact time \textit{\textbf{when}} the Cab will reach my place?"\\ \textcolor{red}{bot} "will be there in 3 minutes."\\ \textcolor{blue}{user} "Yes, that is great, thanks for your work."\\ \textcolor{red}{bot} "Anything else you need?"\\ \textcolor{blue}{user} "No, thanks, that's all I need."\\ \textcolor{red}{bot} "Have a good day."\end{tabular}                                                                                                                                                                                                                                                                                                                                         \\ \bottomrule
\end{tabular}
\end{adjustbox}
\caption{Several examples of extracted dialogue policies for different domains in SGD dataset using the proposed graph-based method, with and without digressions, accompanied by a sample conversation from each domain.}
\label{tab:sample-dialogue-flows}
\end{table*}

\end{document}